\documentclass[10pt,twocolumn,letterpaper]{article}

\usepackage{iccv}
\usepackage{times}
\usepackage{epsfig}
\usepackage{graphicx}
\usepackage{amsmath}
\usepackage{amssymb}
\usepackage{footmisc}

\usepackage[breaklinks=true,bookmarks=false,colorlinks]{hyperref}

\iccvfinalcopy

\begin{document}
\title{MonoLoco: Monocular 3D Pedestrian Localization and Uncertainty Estimation}

\author{Lorenzo Bertoni, Sven Kreiss, Alexandre Alahi\\
EPFL VITA lab\\
CH-1015 Lausanne\\
{\tt\small lorenzo.bertoni@epfl.ch}
}
\maketitle

\begin{abstract}
We tackle the fundamentally ill-posed problem of 3D human localization from monocular RGB images.
Driven by the limitation of neural networks outputting point estimates, we address the ambiguity in the task by predicting confidence intervals through a loss function based on the Laplace distribution. Our architecture is a light-weight feed-forward neural network that predicts 3D locations and corresponding confidence intervals given 2D human poses.
The design is particularly well suited for small training data, cross-dataset generalization, and real-time applications. Our experiments show that we (i) outperform state-of-the-art results on KITTI and nuScenes datasets, (ii) even outperform a stereo-based method for far-away pedestrians,  and (iii) estimate meaningful confidence intervals. We further share insights on our model of uncertainty in cases of limited observations and out-of-distribution samples.
\end{abstract}

\section{Introduction}
Autonomous driving vehicles commonly rely on LiDAR sensing solutions despite high cost and sparsity of point clouds over long ranges \cite{chen2017multi, zhou2018voxelnet, qi2018frustum}.
Cost-effective perception systems have been proposed by adopting stereo/multiple cameras
to address the fundamental ambiguity of monocular solutions \cite{chen20153dop, li2019stereo}. Yet researchers are studying how to push the limits of monocular perception to further contribute to multi-sensor fusion \cite{liang2018deep}.
Progress has been made estimating 3D positions of vehicles from monocular images \cite{ chen2016monocular, mousavian20173d, roddick2018orthographic}, while pedestrians have received far less attention due to lack of adequate performances.
In fact, inferring 3D locations of pedestrians from a single image is particularly ambiguous due to the variance in human heights and shapes.
In this work, we explicitly study the intrinsic ambiguity of locating pedestrians in the scene and investigate whether we can learn this ambiguity from the data.
Driven by this perception task, we aim at providing more insights to the general problem of uncertainty estimation in deep learning.

\begin{figure}
  \centering
  \includegraphics[width=\linewidth]{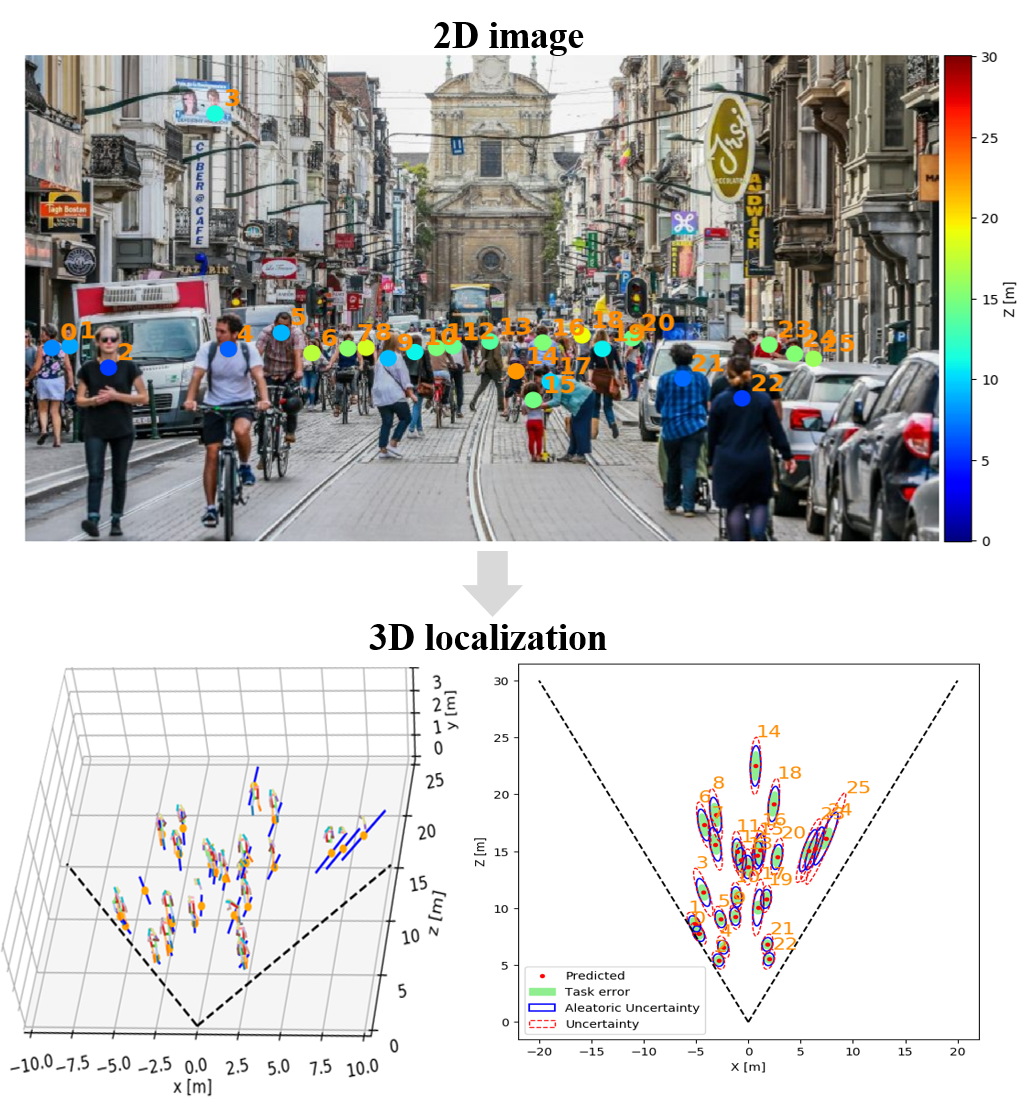}
 \caption{3D localization of pedestrians from a single RGB image. Our method leverages 2D poses to find 3D locations as well as confidence intervals. The confidence intervals are shown as blue lines in the left 3D view and as ellipses in the right birds-eye-view.}
  \label{fig:pull}
\end{figure}

“Kendall and Gal \cite{Kendall2017WhatUD} introduced practical uncertainty estimation for deep learning in perception tasks, distinguishing between \textit{aleatoric} and \textit{epistemic} uncertainty \cite{der2009aleatory, Kendall2017WhatUD}. The former models noise inherent in the observations, while the latter is a property of the model parameters and can be reduced by collecting more data. While their proposed measure of uncertainty is inspiring, they could not compare it with a known uncertainty, referred to as a \textit{task error}. In this work, based on the statistical variation of human height within the adult population \cite{visscher2008sizing}, we quantify the ambiguity of the task, \ie, the task error: an upper bound of performances for monocular 3D pedestrian localization. Surprisingly, the task error is reasonably low. Our experiments show accurate results in 3D localization without overcoming the limitation due to this intrinsic ambiguity.

We propose a simple probabilistic method for monocular 3D localization tailored for pedestrians.
We specifically address the challenges of the ill-posed task by predicting confidence intervals in contrast to point estimates, 
which account for aleatoric and epistemic uncertainties.
Our method is composed of two distinct steps. 
First, we leverage the exceptional progress of pose estimators to obtain 2D joints, a low-dimensional meaningful representation of humans. 
Second, we input the detected joints to a light-weight feed-forward network and output the 3D location of each instance 
along with a confidence interval. 
We explore whether 2D joints contain enough information for a network to learn the intrinsic ambiguity of the task as well as accurate localization.
We leverage a recently introduced loss function based on the Laplace distribution \cite{Kendall2017WhatUD} to incorporate aleatoric uncertainty for each predicted 
location without direct supervision at training time. MC dropout at inference time is used to capture epistemic uncertainty \cite{Gal2016Dropout}.
Our network, referred to as MonoLoco, independently learns the distribution of uncertainties, 
and predicts confidence intervals comparable with the corresponding task error. The code is publicly available online \footnote{\textbf{\url{https://github.com/vita-epfl/monoloco}}\label{code}}.

\section{Related Work}
\paragraph{Monocular 3D Object Detection.}
Recent approaches for monocular 3D object detection in the transportation domain focused only on vehicles as they are rigid objects with known shape. 
To the best of our knowledge, no previous work explicitly evaluated pedestrians from monocular RGB images. 
Kundegorski and Breckon \cite{Kundegorski2014APA} achieved reasonable performances combining infrared imagery and real-time photogrammetry. Alahi \etal combined monocular images with wireless signals \cite{alahi2015rgb} or with additional visual priors \cite{alahi2008object,alahi2014robust}. 
The seminal work of Mono3D \cite{chen2016monocular} exploited deep learning to create 3D object proposals for \textit{car}, \textit{pedestrian} 
and \textit{cyclist} categories but it did not evaluate 3D localization of pedestrians. 
It assumed a fixed ground plane orthogonal to the camera and the proposals were then scored based on scene priors, such as shape, 
semantic and instance segmentations. 
Following methods continued to leverage Convolutional Neural Networks and focused only on \textit{Car} instances. 
To regress 3D pose parameters from 2D detections, Deep3DBox \cite{mousavian20173d}, MonoGRnet \cite{qin2019monogrnet}, and Hu \etal \cite{Hu2018JointM3} used geometrical reasoning for 3D localization, 
while Multi-fusion \cite{xu2018multi} and ROI-10D \cite{manhardt2019roi} incorporated a module for depth estimation. 
Recently, Roddick \etal \cite{roddick2018orthographic} escaped the image domain by mapping image-based features into a birds-eye view representation using integral images. 
Another line of work fits 3D templates of cars to the image~\cite{Xiang2015Datadriven3V, xiang2017subcategory, Chabot2017DeepMA, Kundu20183DRCNNI3}.

While many of the related methods achieved reasonable performances for vehicles, current literature lacks monocular methods addressing other categories in the context of autonomous driving, such as pedestrians and cyclists.

\vspace{-6pt}
\paragraph{Uncertainty in Computer Vision.}
Deep neural networks need to have the ability not only to provide the correct outputs but also a measure of uncertainty, 
especially in safety-critical scenarios like autonomous driving. 
Traditionally, Bayesian Neural Networks~\cite{Richard1991NeuralNC, neal2012bayesian} were used to model epistemic uncertainty 
through probability distributions over the model parameters. However, these distributions are often intractable 
and researchers have proposed interesting solutions to perform approximate Bayesian inference to measure uncertainty,
including Variational Inference \cite{graves2011practical, blundell15weight, salimans2015markov} 
and Deep Ensembles \cite{lakshminarayanan2017simple}. Alternatively, Gal \etal \cite{Gal2016Dropout, gal2017concrete} showed that 
applying dropout \cite{srivastava2014dropout} at inference time yields a form of variational inference where parameters of the network 
are modeled as a mixture of multivariate Gaussian distributions with small variances. 
This technique, called Monte Carlo (MC) dropout, became popular also due to its adaptability to non-probabilistic deep learning frameworks.

In computer vision, uncertainty estimation using MC dropout has been applied for depth regression tasks 
\cite{Kendall2017WhatUD}, scene segmentation \cite{mukhoti2018evaluating, Kendall2017WhatUD} 
and, more recently, LiDAR 3D object detection for cars \cite{feng2018towards}. 

\vspace{-6pt}
\paragraph{Human pose estimation.}
Detecting people in images and estimating their skeleton is a widely studied problem. 
State-of-the-art methods are based on Convolutional Neural Networks and can be grouped into top-down \cite{Papandreou2017TowardsAM, Fang2017RMPERM, He2017MaskR, xiao2018simple} and bottom-up methods  \cite{Cao2017RealtimeM2, newell2017associative, papandreou2018personlab, Kocabas2018MultiPoseNetFM}. 

Related to our work is Simple Baseline~\cite{martinez2017simple}, which showed the effectiveness of latent information contained in 2D joints stimuli. They achieved state-of-the-art results by simply predicting 3D joints from 2D poses through a light, fully connected network. However, similarly to \cite{MorenoNoguer20173DHP, zanfir2018deep, Rogez2019LCRNetM2}, they estimated relative 3D joint positions, not providing any information about the real 3D location in the scene.

\begin{figure}
  \centering
  \includegraphics[width=\linewidth]{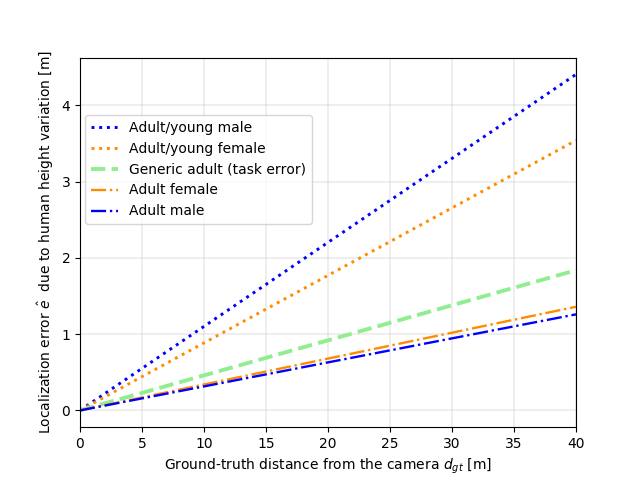}
  \caption{Localization error due to human height variations at different distances from the camera. We approximate the distribution of height for a generic adult as Gaussian mixture distribution and we define the \textit{task error}: an upper bound of performances for monocular methods.}
  \label{fig:task_error}
\end{figure}

\section{Localization Ambiguity}

Inferring depth of pedestrians from monocular images is a fundamentally ill-posed problem. This additional challenge is due to human variation of height. If every pedestrian has the same height, there would be no ambiguity. In this section, we quantify the ambiguity and analyze the maximum accuracy expected from monocular 3D pedestrian localization. 

In our distance estimates, we assume that all humans have the same height $h_\textrm{mean}$ and we analyze the error of this assumption.
Inspired by Kundegorski and Breckon \cite{Kundegorski2014APA}, we model the localization error due to variation of height as a function of the ground truth distance from the camera, which we call \textit{task error}.
From the triangle similarity relation of human heights and distances, $ d_\textrm{h-mean} /h_\textrm{mean} = d_{gt} / h_{gt}$,
where $h_{gt}$ and $d_{gt}$ are the the ground-truth human height and distance, $h_\textrm{mean}$ is the assumed mean height of a person and $d_\textrm{h-mean}$ the estimated distance under the $h_\textrm{mean}$ assumption.
We can define the task error for any person instance in the dataset as:
\begin{equation}
 e \equiv |d_{gt} - d_\textrm{h-mean}| = d_{gt} \; \left| 1 - \frac{h_\textrm{mean}}{h_{gt}} \right| \;\;\;.
\end{equation}
Previous studies from a population of 63,000 European adults have shown that the average height is $178 cm$ for males and $165 cm$ for females 
with a standard deviation of around $7 cm$ in both cases~\cite{visscher2008sizing}. However, a pose detector does not distinguish between genders. 
Assuming that the distribution of human stature follows a Gaussian distribution for male and female populations \cite{freeman1995cross}, 
we define the combined distribution of human heights, a Gaussian mixture distribution $P(H)$, as our unknown ground-truth height distribution. The \textit{expected task error} becomes
\begin{equation}
\hat{e} = d_{gt} \; E_{h \sim P(H) }\left[\left|1 - \frac{h_{mean}}{h}\right|\right] 
 \label{eq:task_error}
\end{equation}
which represents a lower bound for monocular 3D pedestrian localization due to the intrinsic 
ambiguity of the task.
The analysis can be extended beyond adults. A 14-year old male reaches about~$90\%$ of his full height and a female about~$95\%$~\cite{freeman1995cross, Kundegorski2014APA}. 
Including people down to 14 years old leads to an additional source of height variation of~$7.9\%$ and~$5.6\%$ for men and women, respectively~\cite{Kundegorski2014APA}.
Figure~\ref{fig:task_error} shows the expected localization error $\hat{e}$ due to height variations in different cases as a function of the ground-truth distance from the camera~$d_{gt}$. 
This analysis shows that the ill-posed problem of localizing pedestrians, while imposing an intrinsic limit, does not prevent from robust localization in general cases.

\begin{figure*}
  \centering
  \includegraphics[width=\linewidth]{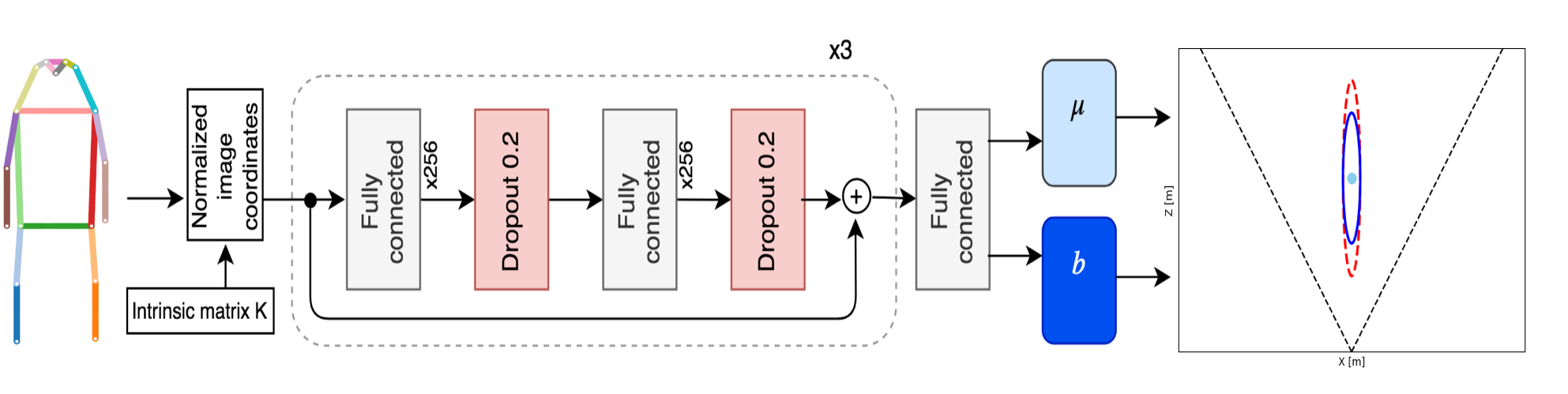}
  \caption{Network architecture. 
  The input is a set of 2D joints extracted from a raw image and the output is the 3D location of a pedestrian $\mu$ and the spread $b$ 
  which represents the associated \textit{aleatoric} uncertainty. The confidence interval is obtained as $\mu \pm b$.
  \textit{Epistemic} uncertainty is obtained through stochastic forward passes applying MC dropout \cite{Gal2016Dropout}. 
  The dashed ellipse represents the two combined uncertainties.
  Every fully connected layer outputs 256 features and is followed by a Batch Normalization layer \cite{ioffe2015batch} and a ReLU activation function.}
  \label{fig:method}
\end{figure*}

\section{Method}
The goal of our method is to detect pedestrians in 3D given a single image.
We argue that effective monocular localization implies not only accurate estimates of the distance but also realistic predictions of uncertainty.
Consequently, we propose a method which learns the ambiguity from the data without supervision and predicts confidence intervals in contrast to point estimates. 
The task error modeled in Eq. \ref{eq:task_error} allows to compare the predicted confidence intervals with the intrinsic ambiguity of the task.

Figure \ref{fig:method} illustrates our overall method, which consists of two main steps. 
First, we exploit a pose detector to escape the image domain and reduce the input dimensionality. 2D joints are a meaningful low-level representation which provides invariance to many factors, including background scenes, lighting, textures and clothes. Second, we use the 2D joints as input to a feed-forward neural network which predicts the distance and the associated ambiguity of each pedestrian. In the training phase, there is no supervision for the localization ambiguity. The network implicitly learns it from the data distribution.

\subsection{Setup}

\paragraph{Input.} 
We use a pose estimator to detect a set of keypoints $\left[u_i, v_i \right]^T $ for every instance in the image. 
We then back-project each keypoint $i$ into normalized image coordinates $ \left[x^*_i, y^*_i, 1 \right]^T $ using 
the camera intrinsic matrix K:

\begin{equation}
\left[x^*_i, y^*_i, 1 \right]^T = K^{-1} \left[u_i, v_i, 1 \right]^T.
\label{eq:k}
\end{equation}

This transformation is essential to prevent the method from overfitting to a specific camera. 
Furthermore, even if we are not predicting a relative 3D location but the distance to the camera, we zero-center the 2D inputs 
around the center coordinates. 
This ensures that the model uses relative distances between joints to make predictions and it prevents overfitting on specific locations 
in the image.

\vspace{-6pt}
\paragraph{2D Human Poses.}
We obtain 2D joint locations of pedestrians using two off-the-shelf pose detectors: the top-down method Mask R-CNN \cite{He2017MaskR}
and the bottom-up one PifPaf~\cite{kreiss2019pifpaf}, both trained on the COCO dataset~\cite{Lin2014MicrosoftCC}. 
The detector can be regarded as a stand-alone module independent from our network, which uses 2D joints as inputs. 
None of the detectors has been fine-tuned on KITTI or nuScenes datasets as no annotations for 2D poses are available.

\vspace{-6pt}
\paragraph{Output.} 
We parametrize the 3D physical location of each instance through its center location $ \textbf{D} = \left[x_c, y_c, z_c \right]^T $. 
We further assume that the projection of the center into the image plane corresponds to the center of the detected bounding box $ \left[u_c, v_c \right]^T $. 
Under these settings, the location of each pedestrian has three degrees of freedom and two constraints. 
We choose to regress the norm of the vector $||\textbf{D}||_2 = \sqrt{x_c^2 + y_c^2 + z_c^2} $ to further constrain 
the location of a pedestrian. For brevity, we will use the notation $ d = ||\textbf{D}||_2$.  
The main criterion is that the dimensions of any object projected into the image plane only depend on the norm of the vector \textbf{D} 
and they are not affected by the combination of its components. 
The same pedestrian in front of a camera or at the margin of the camera field-of-view will appear as having the same height 
in the image plane, as long as the distance from the camera $d$ is the same.

\vspace{-6pt}
\paragraph{Base Network. }
The building blocks of our model are shown in Figure \ref{fig:method}. 
The architecture, inspired by Martinez \etal \cite{martinez2017simple}, is a simple, deep, fully-connected network with six linear layers with 256 output features. 
It includes dropout \cite{srivastava2014dropout} after every fully connected layer, 
batch-normalization \cite{ioffe2015batch} and residual connections \cite{he2016residual}. 
The model contains approximately 400k  training parameters. 

\begin{table*}[tbp!]
 \centering
  \begin{tabular}{|l|c|c c c|c c c|}
    \hline
    Method & Type & & ALP [\%] & & & ALE [m] &\\
      & & ${<0.5m}$&${<1m}$ & ${<2m}$ & $Easy$ & $Moderate$  & $Hard$ \\
    \hline\hline
    Mono3D \cite{chen2016monocular}
    & Mono & 13.2 & 23.2 & 38.9 &  2.13 (2.32) & 2.85 (3.09) &  3.68 (4.46) \\
    MonoDepth \cite{godard2017monodepth} + PifPaf \cite{kreiss2019pifpaf}
    & Mono & 20.5 & 35.3 & 50.6 & 1.48 (1.69) &  2.32 (2.99) & 3.03 (3.67) \\
    Our Geometric baseline
    & Mono & 16.6  & 32.6 & 62.2  & 1.40 (1.48) & 1.35 (1.69) & 1.61 (1.91) \\
    Our MonoLoco - trained on KITTI
    & Mono & 29.0 & 49.6 & 71.2 & 0.94 (0.98) & 1.09 (1.49) & 1.27 (1.90) \\
    Our MonoLoco - trained on nuScenes
    & Mono & \textbf{30.8} & \textbf{51.7} & \textbf{72.1} & \textbf{0.86 (0.92)} & \textbf{1.00 (1.25)} & \textbf{1.17 (1.65)} \\
    \hline\hline
    3DOP \cite{chen20153dop}
    & Stereo & 41.4  & 54.9 & 63.2  & 0.63 (0.71) & 1.18 (1.27) & 1.94 (2.11) \\
    \hline\hline
    Task Error
    & - & 49.0  & 67.3 & 80.0  & 0.62 (0.55) & 0.68 (0.99) & 0.64 (0.75) \\
    \hline
    
  \end{tabular}
  \caption{Comparing our proposed method against baseline results on KITTI dataset ~\cite{Geiger2013Kitti}. 
  The ALE metric is reported for pedestrians commonly detected by all methods to make fair comparison and, on parenthesis, for all the pedestrians detected by each method independently. We outperform all monocular methods and we achieve comparable performances against 3DOP which leverages stereo images for training and testing. Our method uses monocular images and shows cross-dataset generalization when trained on nuScenes dataset \cite{nuscenes}. We use PifPaf \cite{kreiss2019pifpaf} as off-the-shelf network to extract 2D poses.} 
  \label{tab:res_kitti}
\end{table*}

\subsection{Uncertainty}
In this work, we propose a probabilistic network which models two types of uncertainty: \textit{aleatoric} and \textit{epistemic} \cite{der2009aleatory, Kendall2017WhatUD}.

Aleatoric uncertainty is an intrinsic property of the task and the inputs. It does not decrease when collecting more data.
In the context of 3D monocular localization, the intrinsic ambiguity of the task represents
a quota of aleatoric uncertainty. In addition, some inputs may be more noisy than others, leading to an input-dependent aleatoric uncertainty.
Epistemic uncertainty is a property of the model parameters, and it can be reduced by gathering more data. It is useful to quantify the ignorance of the model about the collected data, \textit{e.g.}, in case of out-of-distribution samples.

\vspace{-6pt}
\paragraph{Aleatoric Uncertainty.}

Aleatoric uncertainty is captured through a probability distribution over the model outputs.
We define a relative Laplace loss based on the negative log-likelihood of a Laplace distribution as:
\begin{equation}
  L_{\textrm{Laplace}}(x|\mu,b) = \frac{|1-\mu/x|}{b} + \log(2b)
\label{eq:laplace} 
\end{equation}
where $x$ is the ground truth and $\{\mu,b\}$ are the parameters
predicted by the model. 
$\mu$ represents the predicted distance while $b$ is the spread, making this training objective an attenuated $L_1$-type loss via spread $b$. 
During training, the model has the freedom to predict a large spread $b$, leading to attenuated gradients for noisy data.
At inference time, the model predicts the distance $\mu$ and a spread $b$ 
which indicates its confidence about the predicted distance.
Following~\cite{Kendall2017WhatUD}, to avoid the singularity for $b=0$, 
we apply a change of variable to predict the log of the spread $s = \log(b)$.

Compared to previous methods \cite{Kendall2017WhatUD, wirges2019capturing}, 
we design a Laplace loss which works with relative distances to keep into account the role of distance in our predictions. 
Estimating the distance of a pedestrian with an absolute error can lead to a fatal accident if the person is very close, 
or be negligible if the same human is far away from the camera. 

\vspace{-6pt}
\paragraph{Epistemic Uncertainty.}
To model epistemic uncertainty, we follow \cite{Gal2016Dropout, Kendall2017WhatUD} and consider each parameter 
as a mixture of two multivariate Gaussians with small variances and means $0$ and $\theta$. 
The additional minimization objective for N data points is:
\begin{equation}
  L_{\textrm{dropout}}(\theta, p_{drop}) = \frac{1-p_{drop}}{2N}||\theta||^2 \;\;\; .
 \label{mc_drop}
\end{equation}

In practice, we perform dropout variational inference by training the model with dropout before every weight layer 
and then performing a series of stochastic forward passes at test time using the same dropout probability $p_{drop}$ of training time. 
The use of fully-connected layers makes the network particularly suitable for this approach, 
which does not require any substantial modification of the model.

The combined epistemic and aleatoric uncertainties are captured by the sample variance of predicted distances $\tilde{x}$. They are sampled from multiple Laplace distributions parameterized with the predictive distance $\mu$ and spread $b$ from multiple forward passes with MC dropout:
\begin{align}
  Var(\tilde{X}) =
  & \frac{1}{TI} \sum_{t=1}^T \sum_{i=1}^I \tilde{x}_{t,i}^2(\mu_t, b_t) \nonumber \\
  & - \left[ \frac{1}{TI} \sum_{t=1}^T \sum_{i=1}^I \tilde{x}_{t,i}(\mu_t, b_t) \right]^2
\label{eq:variance}
\end{align}
where for each of the $T$ computationally expensive forward passes, $I$ computationally cheap samples are drawn from the Laplace distribution.

\section{Experiments}

\subsection{Implementation details.}
\paragraph{Datasets. }
We train and evaluate our model on KITTI Dataset ~\cite{Geiger2013Kitti}. It contains 7481 training images along with camera calibration files. All the images are captured in the same city from the same camera. To analyze cross-dataset generalization properties, we train another model on the teaser of the recently released \emph{nuScenes} dataset~\cite{nuscenes} and we test it on KITTI. We do not perform cross-dataset training.

\vspace{-6pt}
\paragraph{Training/Evaluation Procedure. }
To obtain input-output pairs of 2D joints and distances, we apply an off-the-shelf pose detector and use intersection over union of 0.3 to match our detections with the ground-truths, obtaining 5000 instances for KITTI and 14500 for nuScenes teaser. KITTI images are upscaled by a factor of two to match the minimum dimension of 32 pixels of COCO instances. NuScenes already contains high-definition images, which are not modified.
We follow the KITTI train/val split of Chen \etal \cite{chen2016monocular} and we run the training procedure for 200 epochs using Adam optimizer~\cite{kingma2014adam}, a learning rate of $10^{-3}$ and mini-batches of 512. 
The code, available online \footref{code}, is developed using PyTorch~\cite{pytorch}. 
Working with a low-dimensional latent representation is very appealing as it allows fast experiments with different architectures and hyperparameters.

\subsection{Evaluation.}
\paragraph{Localization Error.}
We evaluate 3D pedestrian localization using the Average Localization Precision (ALP) metric 
defined by Xiang \etal \cite{Xiang2015Datadriven3V} for the \textit{car} category. 
ALP considers a prediction as correct if the error between the predicted distance and the ground-truth is smaller 
than a threshold. 
We also analyze the average localization error (ALE) in two different conditions.
Following KITTI guidelines, we split the instances in three difficulty regimes based on bounding box height, levels of occlusion and truncation: \textit{easy}, \textit{medium} and \textit{hard}. 
We also compare the results against the task error of Eq.~\ref{eq:task_error}, which defines the target error for monocular approaches due to the ambiguity of the task. 

\begin{figure}
  \centering
  \includegraphics[width=\linewidth]{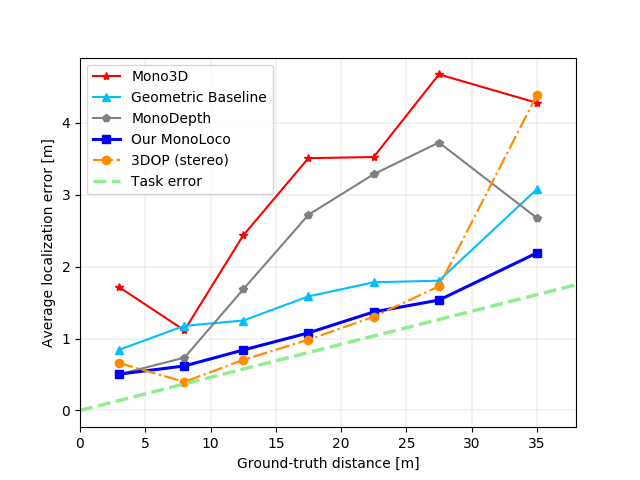}
  \caption{Average localization error for the instances commonly detected by all methods. 
  We outperform the monocular Mono3D \cite{chen2016monocular} while achieving comparable performances 
  with the stereo 3DOP \cite{chen20153dop}. Monocular performances are bounded by our modeled task error
  in Eq. \ref{eq:task_error}.}
  \label{fig:results}
\end{figure}

\vspace{-6pt}
\paragraph{Geometrical Approach.}
3D pedestrian localization is an ill-posed task due to human height variations.  
On the other side, estimating the distance of an object of known dimensions from its projections 
into the image plane is a well-known deterministic problem. As a baseline, we consider humans as fixed objects 
with the same height and we investigate the localization accuracy under this assumption.

For every pedestrian, we apply a pose detector to calculate distances in pixels between different body parts in the image domain. 
Combining this information with the location of the person in the world domain, we analyze the distribution of the real dimensions (in meters) of all the instances in the training set for three segments: 
head to shoulder, shoulder to hip and hip to ankle.

For our calculation we assume a pinhole model of the camera and that all instances stand upright.
Using the camera intrinsic matrix K and knowing the ground truth location of each instance 
$ \textbf{D} = \left[x_c, y_c, z_c \right]^T $ we can back-project each keypoint from the image plane to its 3D 
location and measure the height of each segment using Eq. \ref{eq:k}. 
We calculate the mean and the standard deviation in meters of each of the segments for all the instances in the training set. 
The standard deviation is used to choose the most stable segment for our calculations. 
For instance, the position of the head with respect to shoulders may vary a lot for each instance. 
To take into account noise in the 2D joints predictions we also average between left and right keypoints values. 
The result is a single height $\Delta y_{1-2}$ which represents the average length of two body parts. 
In practice, our geometric baseline uses the \textit{shoulder-hip} segment and predicts an average height of $50.5cm$. Combining the study on human heights \cite{visscher2008sizing} described in Section 3  with the anthropometry study of Drillis \etal \cite{drillis1969body}, we can compare our estimated $\Delta y_{1-2}$ with the human average \textit{shoulder-hip} height: $0.288 * 171.5cm = 49.3cm$.

The next step is to calculate the location of each instance knowing the value in pixels of the chosen keypoints  $v_1$ and $v_2$ and assuming $\Delta y_{1-2}$ to be their relative distance in meters. 
This configuration requires to solve an over-constrained linear system with two specular solutions, of which only one is inside the camera field of view.

\begin{figure}
  \centering
  \includegraphics[width=\linewidth]{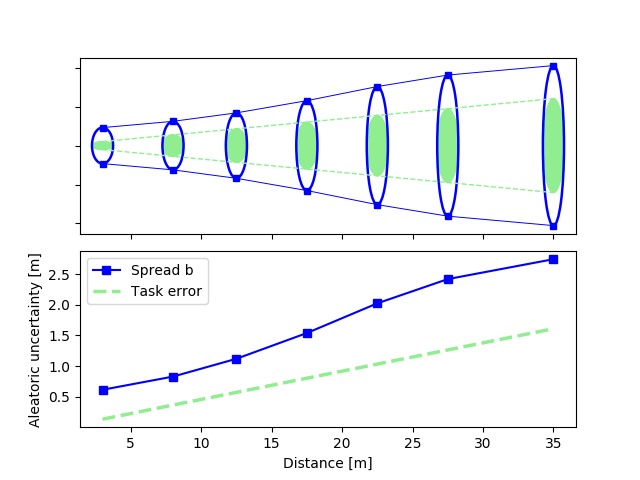}
  \caption{Results of aleatoric uncertainty predicted by MonoLoco (spread $b$) 
  and the modeled aleatoric uncertainty due to human height variation (task error $\hat{e}$). 
  The term $b - \hat{e}$ is indicative of the aleatoric uncertainty due to noisy observations. 
  On the top figure, we visualize the average predicted and ground truth confidence intervals $\pm b$ and $ \pm \hat{e}$  at various distances, using ellipses with minor axis of one meter as a reference.} 
  \label{fig:spread}
\end{figure}

\vspace{-6pt}
\paragraph{Baselines.}
We compare our method on KITTI against two monocular approaches and a stereo one:
\begin{itemize}
\vspace{-5pt}
   \item \textbf{Mono3D} \cite{chen2016monocular} is a monocular 3D object detector for cars, cyclists and pedestrians. 3D localization of pedestrians is not evaluated but detection results are publicly available. 
   \vspace{-5pt}
   \item \textbf{MonoDepth} \cite{godard2017monodepth} is a monocular depth estimator which predicts a depth value for each pixel in the image. To estimate a reference depth value for every pedestrian, we detect 2D joints using PifPaf and calculate the depth for a set of 9 pixels around each keypoint. We then consider the minimum depth as our reference value. Experimentally, the minimum depth increases the performances compared to the average one. From the depth, we calculate the distance $d$ using the normalized image coordinates of the center of the bounding box.
  \vspace{-5pt}
   \item \textbf{3DOP} \cite{chen20153dop} is a stereo approach for pedestrians, cars and cyclists and their 3D detections are publicly available.
\end{itemize}

\begin{figure}
  \centering
  \includegraphics[width=\linewidth, height=3.1cm]{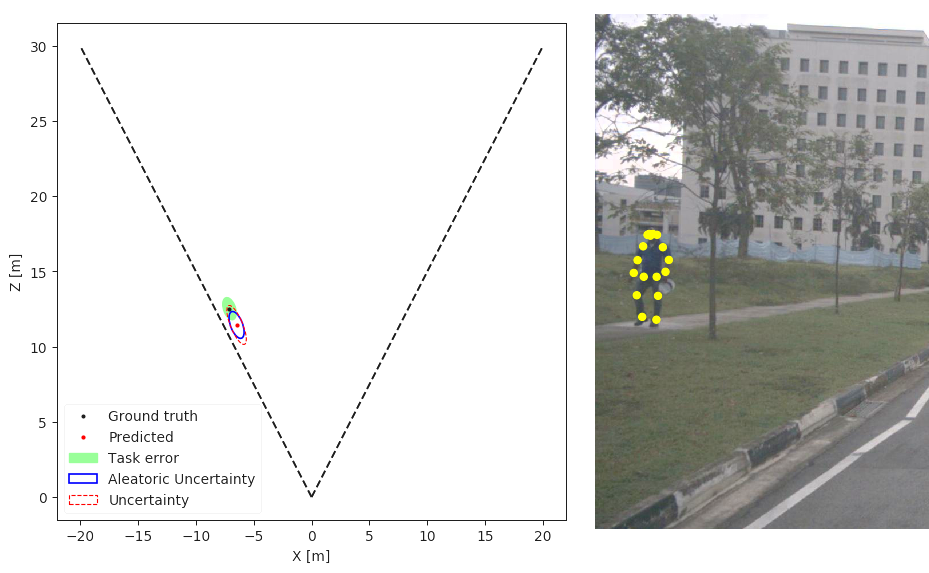}
  \includegraphics[width=\linewidth, height=3.1cm]{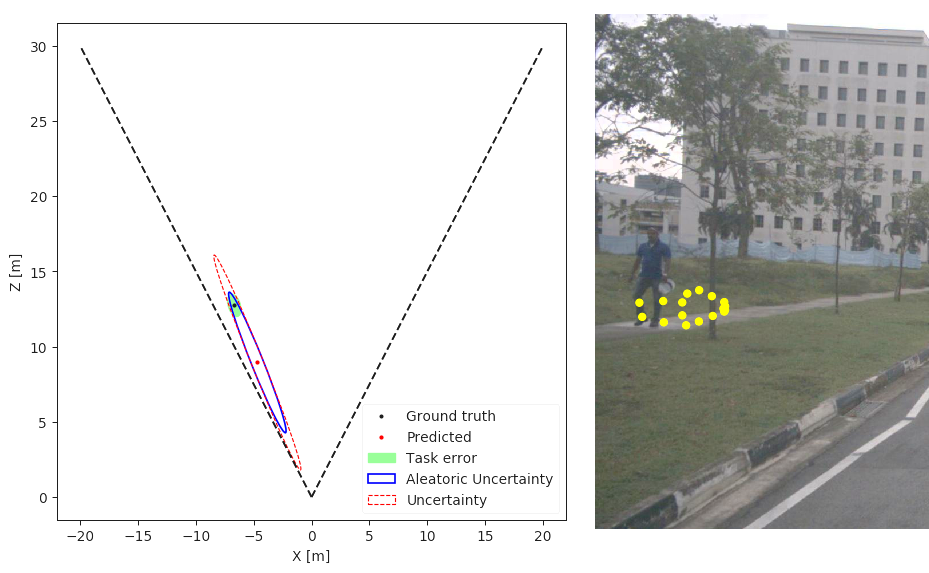}
  \caption{Simulating the outlier case of a person lying on the ground. 
  In the top image, the predicted confidence interval is small and the detection accurate. In the bottom image, we create an outlier pose by projecting on the ground the original pose. The network predicts higher uncertainty, a useful indicator to warn about out-of-distribution samples.}
  \label{fig:outliers}
\end{figure}

\begin{table}[tbp]
  \centering
  \begin{tabular}{|l| c c c|}
    \hline
    & $|x- \mu|/ \sigma$ &$ |\sigma - e|$ [m] & Recall [\%]\\
    \hline
    \hline
    $p_{drop}=0.05 $  & 0.60 & 0.90 & 82.8  \\
    $p_{drop}=0.2 $  & 0.58 & 0.96 & 84.3 \\
    $p_{drop}=0.4 $  & 0.50 & 1.26 & 88.3 \\
    \hline
  \end{tabular}
  \caption{Precision and recall of uncertainty for KITTI validation set with 50 stochastic forward passes. $|x- \mu|$ is the localization error, $\sigma$ the predicted confidence interval, $\hat{e}$ the task error modeled in Eq. \ref{eq:task_error} and Recall is represented by the \% of ground truth instances inside the predicted confidence interval.}
  \label{tab:uncertainty}
\end{table}

\begin{table}[tbp!]
  \centering
  \begin{tabular}{|l|c c c c c|}
    \hline
    
    Mask R-CNN  & & & ALE [m] & & \\
    \cite{He2017MaskR} & ${_0^{10}}$ & ${_{10}^{20}}$ & ${_{20}^{30}}$ & ${_{30}^+}$ & ${All}$ \\
    \hline\hline
    Geometric
    & 0.79 & 1.52 & 3.17 & 9.08 & 3.73 \\
    \hline
    $L_1$ loss
    & 0.85 & \textbf{1.17} & 2.24 & 4.11 & 2.14 \\
    Gaussian loss
    & 0.90 & 1.28 & 2.34 & 4.32 & 2.26  \\
    Laplace Loss
    & \textbf{0.74} & \textbf{1.17} & 2.25 & 4.12 & 2.12  \\
    \hline
    \hline
    PifPaf \cite{kreiss2019pifpaf} & & & ALE [m] & & \\
    & ${_0^{10}}$ & ${_{10}^{20}}$ & ${_{20}^{30}}$ & ${_{30}^+}$ & ${All}$ \\
    \hline\hline
    Geometric
    & 0.83 & 1.40 & 2.15 & 3.59 & 2.05 \\
    \hline
    $L_1$ loss
    & 0.83 & 1.24 & \textbf{2.09} & 3.32 & 1.92 \\
    Gaussian loss
    & 0.89 & 1.22 & 2.14 & 3.50 & 1.97 \\
    \textbf{Laplace Loss}
    & 0.75 & 1.19 & 2.24 & \textbf{3.25} & \textbf{1.90} \\
    \hline
  \end{tabular}
  \caption{Impact of different loss functions and pose detectors on nuScenes teaser validation set \cite{nuscenes}.}
  \label{tab:ablation}
\end{table}

\begin{table}[tbp!]
  \centering
  \begin{tabular}{|l| c c c|}
    \hline
    Method $ \setminus $ Time [ms] & $t^{pose}$ & $t^{model}$& $t^{total}$\\
    \hline
    \hline
    Mono3D \cite{chen2016monocular} & - & 1800 & 1800   \\
    3DOP \cite{chen20153dop} & - & $~$2000 & $~$2000 \\
    Our (1 forward pass) & 89 / 162 & 10  & \textbf{99 / 172} \\
    Our (50 forward passes) & 89 / 162 & 51 & 140 / 213\\
    \hline
  \end{tabular}
  \caption{ Single-image inference time on a GTX 1080Ti for KITTI dataset with Pifpaf as pose detector. We only considered images with positive detections. Most computation comes from the pose detector (ResNet 50 / ResNet 152 backbones). For Mono3D and 3DOP we report published statistics on a Titan X GPU.
 }
  \label{tab:runtime}
\end{table}

\begin{figure*}
  \centering
      \vspace{-20pt}
    \includegraphics[width=0.9\linewidth, height=5.0cm]{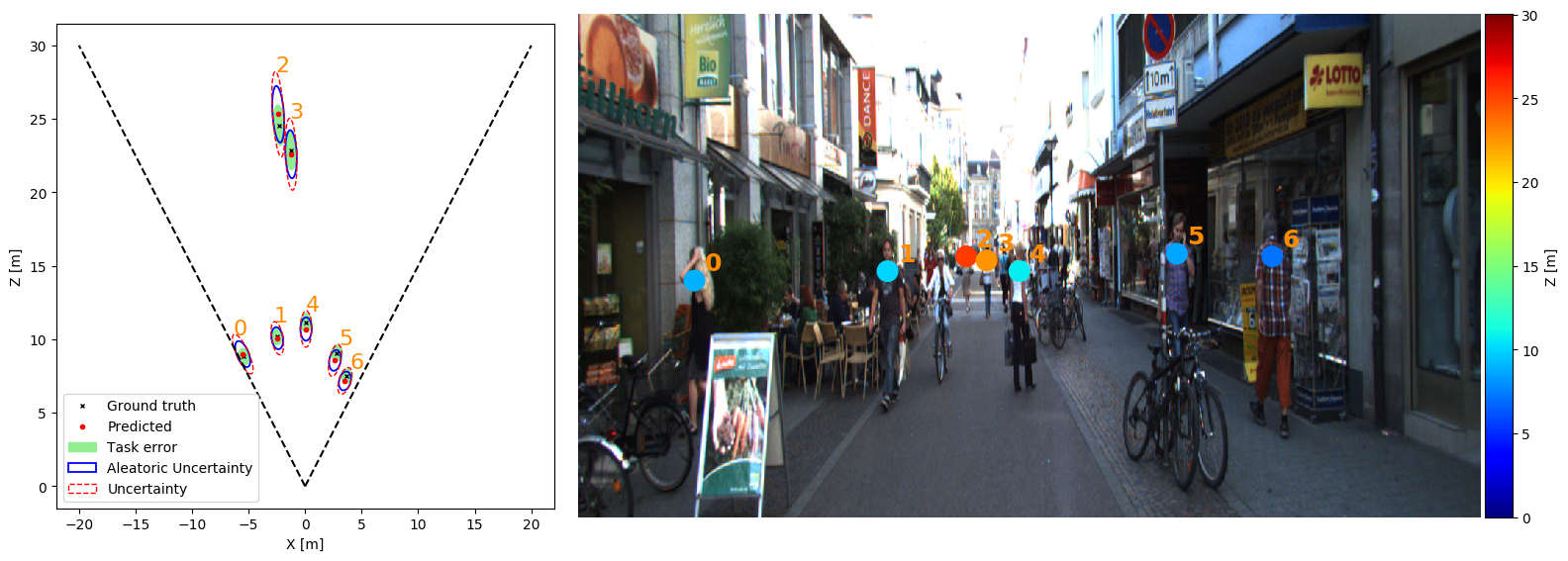}
    \vspace{-3pt}
    \includegraphics[width=0.9\linewidth, height=5.0cm]{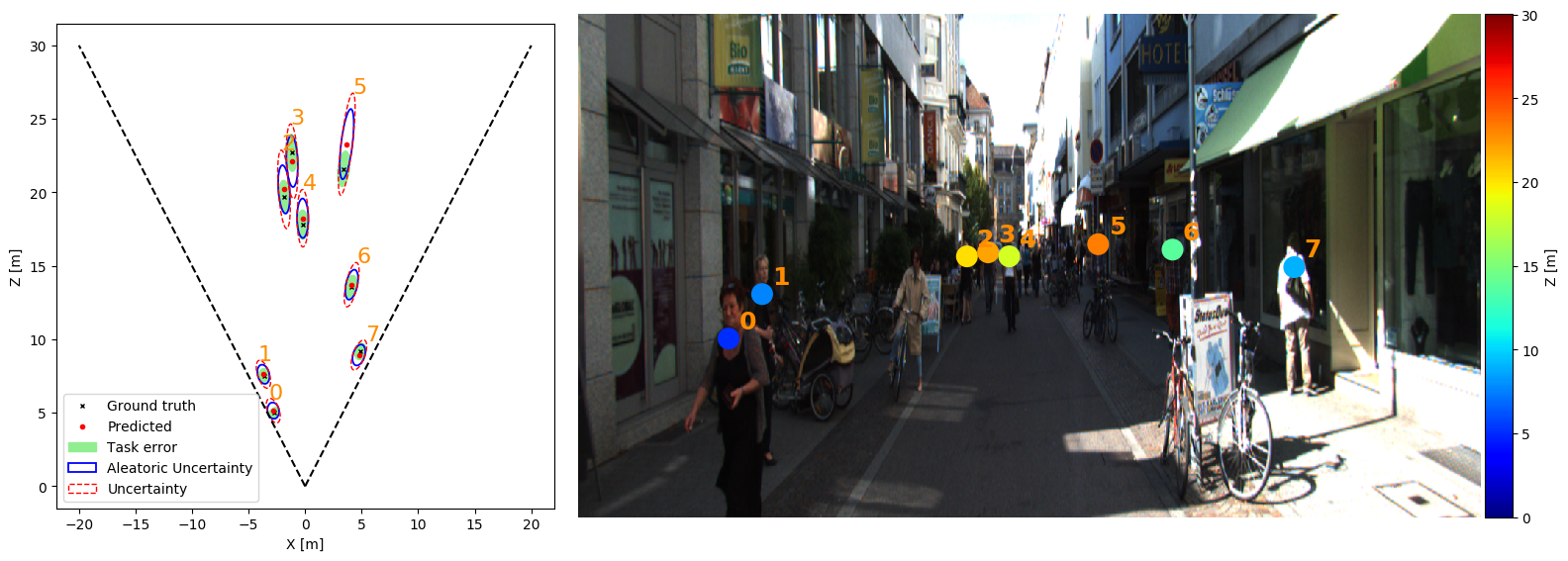}
    \includegraphics[width=0.9\linewidth, height=5.0cm]{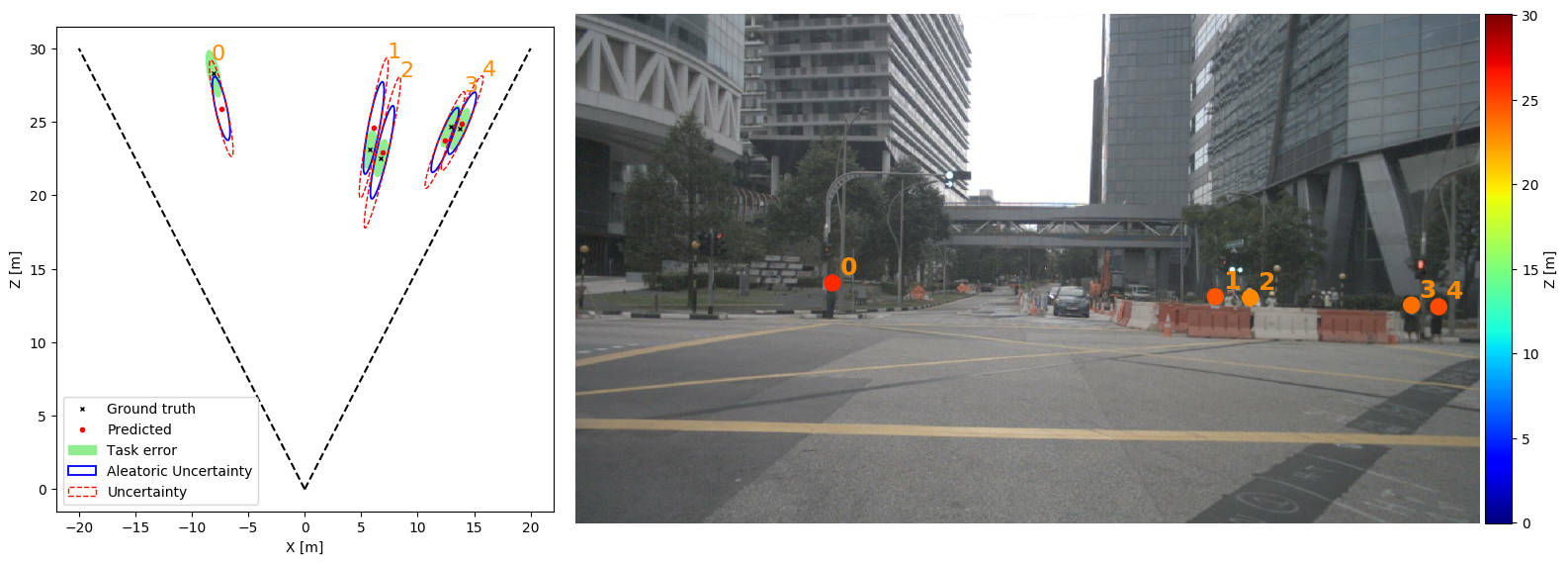}

    \caption{Illustration of results from KITTI ~\cite{Geiger2013Kitti} (top and middle) and nuScenes ~\cite{nuscenes} (bottom)  datasets containing true and inferred distance information as well as confidence intervals (represented by ellipses with minor axis of one meter). We observe that the predicted uncertainty increases in case of occlusions (bottom image, pedestrians 1 and 2).}
  \label{fig:qual}
\end{figure*}

\subsection{Results. }
\paragraph{Localization Accuracy.} 

Table \ref{tab:res_kitti} summarizes our quantitative results on KITTI. 
We strongly outperform all the other monocular approaches on all metrics with any of the two models trained either on KITTI or nuScenes. 
We obtain comparable results with the stereo approach 3DOP  \cite{chen20153dop}, which
has been trained and evaluated on KITTI 
and makes use of stereo images during training and test time. 

In Figure \ref{fig:results}, we make an in-depth comparison analyzing the average localization error as a function of the ground truth distance, while Figure \ref{fig:qual} shows qualitative results on challenging images from KITTI and nuScenes datasets.A video with additional results is available online. \footnote{\textbf{\url{https://youtu.be/ii0fqerQrec}}\label{video}}.

\vspace{-6pt}
\paragraph{Aleatoric Uncertainty. }
We compare in Figure \ref{fig:spread} the aleatoric uncertainty predicted by our network through spread \textit{b} with the \textit{task error} due to human height variation defined in Eq. \ref{eq:task_error}.
The predicted spread $b$ is a property of each set of inputs and, differently from $\hat{e}$, is not only a function of the distance 
from the camera $d$. Indeed, the predicted aleatoric uncertainty includes  not only the uncertainty due to the ambiguity of the task, but also the uncertainty due to noisy observations \cite{Kendall2017WhatUD}, \ie, the 2D joints inferred by the pose detector.
Hence, we can approximately define the predictive aleatoric uncertainty due to noisy joints as $b - \hat{e}$ 
and we observe that the further a person is from the camera, the higher is the term $b - \hat{e}$.
The spread $b$ is the result of a probabilistic interpretation of the model and the resulting confidence intervals are calibrated. On KITTI validation set they include 68\% of the instances. 

\vspace{-6pt}
\paragraph{Combined Uncertainty. }
The combined aleatoric and epistemic uncertainties are captured by sampling from multiple Laplace distributions using MC dropout. The magnitude of the uncertainty depends on the chosen dropout probability $p_{\textrm{drop}}$ in Eq. \ref{mc_drop}. 
In Table \ref{tab:uncertainty}, we analyze the precision/recall trade-off for different dropout probabilities and choose $p_{\textrm{drop}} = 0.2$. We perform 50 computationally expensive forward passes and, for each of them, 100 computationally cheap samples from Laplace distribution using Eq.~\ref{eq:variance}. As a result, $84\%$ of pedestrians lie inside the predicted confidence intervals for the validation set of KITTI.

Our final goal is to make self-driving cars safe and being able to predict a confidence interval instead of a single regression number is a first step towards this direction. 
To illustrate the benefits of predicting intervals over point estimates, we construct a controlled risk analysis. 
We define as \textit{high-risk cases} all those instances where the ground truth distance is smaller than the predicted one, 
hence a collision is more likely to happen. 
We estimate that among the 1932 detected pedestrians in KITTI which match a ground truth, 48\% of them are considered as \textit{high-risk cases}, 
but for 89\% of them the ground truth lies inside the predicted interval. 

\vspace{-6pt}
\paragraph{Outliers.}
Leveraging on the simplicity of manipulation of 2D joints, we analyze the role of the predicted uncertainty in case of an outlier. As shown in Figure \ref{fig:outliers}, we recreate the pose of a person lying down and we compare it with a ``standard" detection of the same person standing up. When the pedestrian is lying down, the network predicts an unusually large confidence interval which includes the ground truth location.

In the bottom image of Figure \ref{fig:qual}, we also highlight the behavior of the model in case of partially occluded pedestrians (pedestrians 1 and 2), where we also empirically observe larger confidence intervals when compared to visible pedestrians at similar distances.

\vspace{-6pt}
\paragraph{Ablation studies.}
In Table \ref{tab:ablation} we analyze the effects of choosing a top-down or a bottom-up pose detector with different loss functions and with our deterministic geometric baseline.
$L_1$-type losses perform slightly better than the Gaussian loss, but the main improvement is given by choosing PifPaf as pose detector.

\vspace{-6pt}
\paragraph{Run time.}
A run time comparison is shown in Table \ref{tab:runtime}. Our method is 9-20 times faster than compared methods (depending on the pose detector backbone) and it is the only one suitable for real-time applications.
\vspace{-0.5pt}
\section{Conclusions}
\vspace{-2pt}
We have proposed a new approach for 3D pedestrian localization based on monocular images which addresses the intrinsic ambiguity of the task by predicting calibrated confidence intervals.
We have shown that our method even outperforms a stereo approach at further distances 
because it is less sensitive to low-resolution imaging issues. 

For autonomous driving applications, combining our method with a stereo approach is an exciting direction for accurate, low-cost and real-time 3D localization.

\textbf{Acknowledgements} We acknowledge the support of
Samsung and  Farshid Moussavi for helpful discussions.

{\small
\bibliographystyle{ieee_fullname}
\bibliography{references}
}

\end{document}